\documentclass[11pt]{article}

\usepackage[preprint]{acl}

\usepackage{times}
\usepackage{latexsym}

\usepackage[T1]{fontenc}

\usepackage[utf8]{inputenc}

\usepackage{microtype}

\usepackage{inconsolata}

\usepackage{graphicx}

\usepackage{hyperref}
\usepackage{url}
\usepackage{multirow}
\usepackage{tabularx}
\usepackage{booktabs}
\usepackage{subcaption}
\usepackage{wrapfig}
\usepackage{xcolor,colortbl}
\usepackage{amsmath}
\usepackage{amsfonts}

\title{MCA: Modality Composition Awareness for Robust \\ Composed Multimodal Retrieval}

\usepackage{soul}
\soulregister\citep7
\soulregister\citet7
\soulregister\ref7
\soulregister\eqref7

\sethlcolor{yellow!30}

\newif\ifshowrevisions
\showrevisionsfalse   

\newcommand{\rev}[1]{%
  \ifshowrevisions
    \texorpdfstring{\hl{#1}}{#1}
  \else
    #1
  \fi
}

\newcommand{\ourmethod}{MCA}

\definecolor{Gray}{gray}{0.85}
\newcolumntype{a}{>{\columncolor{Gray}}c}
\newcolumntype{Y}{>{\centering\arraybackslash}X}

\author{
\textbf{Qiyu Wu}\thanks{Corresponding to: qiyu.wu@sony.com.},
\textbf{Shuyang Cui},
\textbf{Satoshi Hayakawa}\thanks{Currently at The University of Tokyo. Work done while at Sony Group Corporation.},
\textbf{Wei-Yao Wang}, \\
\textbf{Hiromi Wakaki},
\textbf{Yuki Mitsufuji}
\\
Sony Group Corporation
\\
\texttt{\{firstname.lastname\}@sony.com}
}

\begin{document}

\maketitle

\begin{abstract}
Multimodal retrieval, which seeks to retrieve relevant content across modalities such as text or image, supports applications from AI search to contents production. Despite the success of separate-encoder approaches like CLIP aligning modality-specific embeddings with contrastive learning, recent multimodal large language models (MLLMs) enable a unified encoder that directly processes composed inputs. While flexible and advanced, we identify that unified encoders trained with conventional contrastive learning are prone to learn modality shortcut, leading to poor robustness under distribution shifts. We propose a modality composition awareness framework to mitigate this issue. Concretely, it consists of a preference loss enforces multimodal embeddings to outperform their unimodal counterparts, and a composition regularization objective aligns multimodal embeddings with prototypes composed from its unimodal parts. These objectives explicitly model structural relationships between the composed representation and its unimodal counterparts. Experiments on various benchmarks show gains in out-of-distribution retrieval, highlighting modality composition awareness as a effective principle for robust composed multimodal retrieval when utilizing MLLMs as the unified encoder.
\end{abstract}

\section{Introduction}
\label{sec:intro}
Multimodal retrieval, which aims to retrieve semantically relevant contents across multiple modalities such as text, image and audio, is a fundamental task in various information fields. Its applications span a wide range of domains, such as
text-vision retrieval~\citep{huynh2025collm, wang2021t2vlad}, music retrieval~\citep{doh2023toward}, product search~\citep{goenka2022fashionvlp,zhu2024bringing},
and multimodal retrieval-augmented generation~\citep{yasunaga2023retrieval,ghosh2024recap,yang2024audiobox,jeong2025videorag}.
The core ability of multimodal retrieval is to represent multimodal inputs in a shared and comparable embedding space.
A prevailing approach to this problem is to adopt unimodal encoders and align the encoded embeddings through contrastive learning (CL). Models following this separate-encoder paradigm, such as CLIP~\citep{radford2021learning} and CLAP~\citep{elizalde2023clap}, have demonstrated the effectiveness of CL, achieving strong performance across various multimodal retrieval tasks.
On the other hand, with the rapid development of multimodal large language models (MLLMs)~\citep{alayrac2022flamingo,li2023blip,bai2023qwen,chu2023qwen,liu2023visual, chen2024internvl}, there has been growing interest in employing MLLMs as encoders for multimodal retrieval~\citep{jiang2024e5,zhang2024gme,huang2025joint,jiang2025vlmvec}.
Unlike separate-encoder frameworks, MLLMs are capable of processing inputs from different modalities, as well as their compositions, within a unified architecture, providing several advantages such as a shared semantic space across modalities, flexible handling of composed queries and documents, e.g., text+image, and the ability to leverage powerful pretrained representations including rich language knowledge and multimodal understanding.

\begin{figure}[ht]
    \centering
    \includegraphics[width=\linewidth]{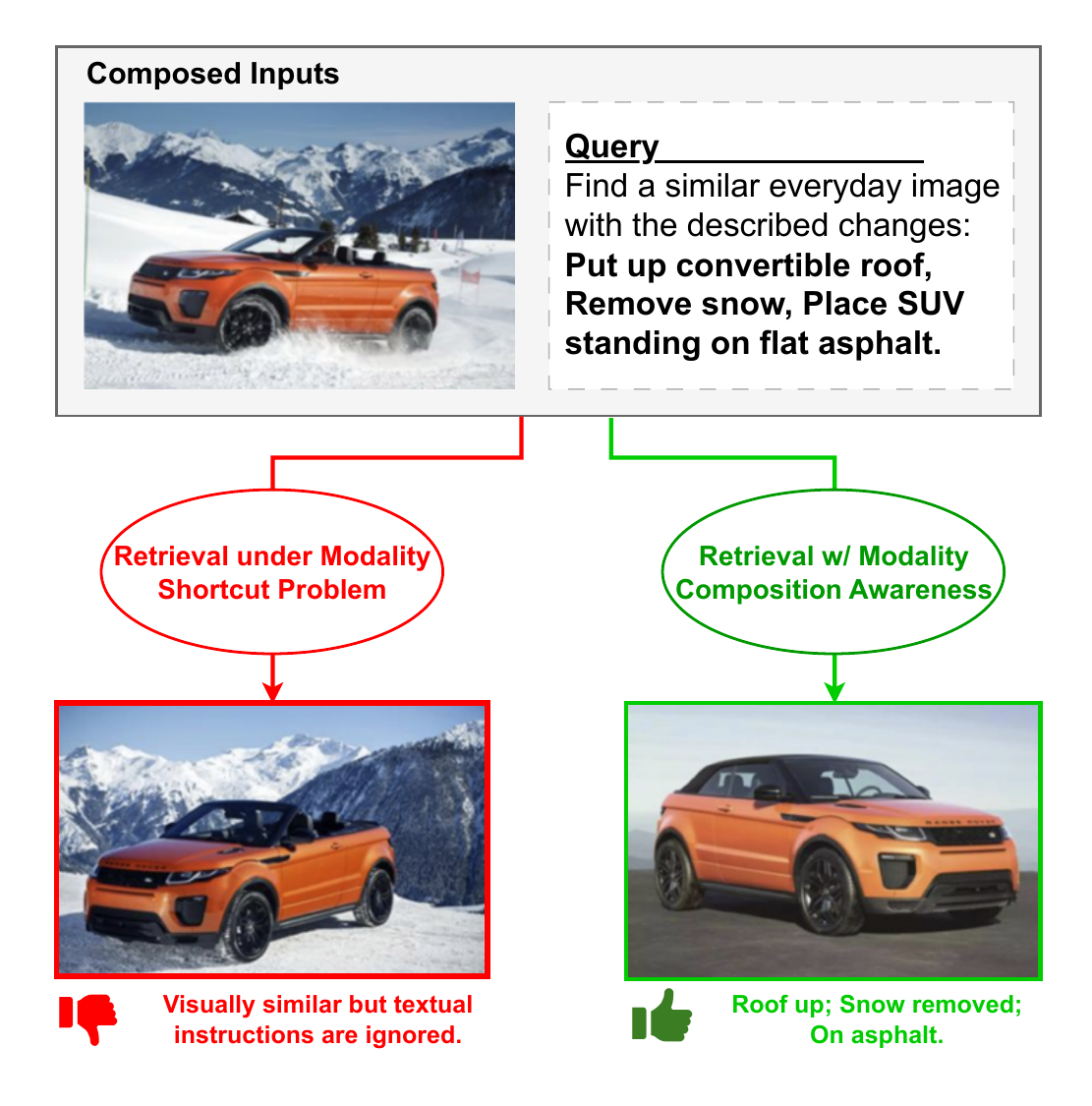}
    \caption{Illustration of modality shortcut problem in multimodal retrieval. Given the inputs consists of an image and textual instruction, \textbf{left}: baseline model trained with conventional CL loss retrieve a visually similar image, focusing too much on vision modality; \textbf{right}: model trained with modality composition awareness retrieves the relevant image following the instruction.}
    \label{fig:fig1}
\end{figure}

However, this flexibility also makes the model more prone to \textbf{modality shortcut learning}. Since all modalities are jointly processed within a shared encoder, the training loss can be minimized by over-relying on the stronger modality signal, while ignoring the complementary one.
An example in Figure~\ref{fig:fig1} illustrates that the model suffers from modality shortcuts when processing highly similar images, while ignoring the textual instruction:
``put up convertible roof, remove snow, place SUV standing on flat asphalt''.
Hence, the architecture shift from separate-encoder to unified-encoder makes the direct application of conventional CL objective to unified MLLM encoders limiting to robustness.
Furthermore, the modality shortcut issue naturally extends to scenarios involving multiple modalities, particularly as recent work aims to unify various modalities into a single framework~\citep{girdhar2023imagebind,zhu2024languagebind,xu2025qwen2}. As the number of modalities increases, so does the risk of the model collapsing to rely on a single dominant modality, making explicit modality composition-aware constraints crucial for robust generalization.

We argue that \textbf{modality composition} is the key to mitigate this issue. In this paper, we propose a \emph{modality composition awareness} (\ourmethod{}) framework that modeling the structural relationship between multimodal and unimodal representations, as shown in Figure~\ref{fig:approach}. \ourmethod{} consists of two complementary objectives from the preference and consistency perspective, respectively. First, a \emph{preference loss} enforces that the embeddings of a multimodal composition should be more discriminative than any of its unimodal counterparts, thereby discouraging modality shortcut learning. Second, a \emph{composition regularization} objective encourages the consistency between the composed embedding produced by the unified encoder and a compositional prototype constructed from its unimodel embeddings, ensuring that composed representations remain grounded in their constituent modalities. Together, these objectives explicitly model the structural relationship between multimodal and unimodal inputs, leading to more robust representation. Figure~\ref{fig:tsneours} visualizes the representation learned with \ourmethod{}. It demonstrates that composed queries, unimodal queries and targets have a clearer boundary, showing that \ourmethod{} reduces shortcut reliance.

To comprehensive evaluate \ourmethod{}, we conduct extensive experiments on both in-domain (IND) and OOD benchmarks, covering retrieval and grounding tasks. The results demonstrate the \ourmethod{} improves robustness with OOD improvements under distribution shifts while maintaining IND performance. Through ablations, we show that both the preference and regularization components contribute complementary benefits. In addition, we also highlight the interaction between input richness and the strength of our proposed losses. Those results suggest that explicitly modeling modality composition could be a general principle with broader implications for the unified multimodal retrieval using MLLMs.
\section{Related Work}
\paragraph{Multimodal Retrieval}
Classical multimodal retrieval typically adopts separate-encoder architectures, learning a text encoder and an image/audio/video encoder whose outputs are aligned in a share space via CL loss. This line of work has achieved strong performance at scale and efficient retrieval~\citep{radford2021learning,li2022blip,zhai2023sigmoid}.
Beyond unimodal inputs, composed retrieval focuses inputs with multiple modalities, such as text+image or text+audio, which better capture user intent in real-world scenarios. This setting has been studied in contexts like text-guided image retrieval~\citep{yu2016modeling}, fashion production search~\citep{wu2021fashion} and multimodal modification tasks~\citep{liu2021image}.
Separate-encoder approaches are principally designed for unimodal queries. When the query or candidate is a composition of modalities, an extra fusion module~\citep{wei2024uniir,huynh2025collm} has to be trained to fuse the modalities. While these methods demonstrate the feasibility of handling composed queries, they are typically build upon conventional CL loss and do not model the cross-modal interaction between the modalities, leading to a performance sacrifice~\citep{huang2025joint}.
\rev{
Specifically, UniIR~\citep{wei2024uniir} fuses unimodal embeddings to enable dual-encoders for composed retrieval, while CoLLM~\citep{huynh2025collm} synthesizes composed inputs to augment training data. In contrast, MCA constructs compositional prototypes from unimodal parts within each sample as regularization, specifically targeting the modality shortcut problem in unified MLLM encoders, and this issue is not discussed by prior works.
}

\paragraph{MLLMs for Retrieval}
Recent MLLMs extend pretrained LLMs with multimodal adapters or cross-modal modules, enabling an integrated architecture to process and understand over text images, audios and videos~\citep{alayrac2022flamingo,li2023blip,bai2023qwen,chu2023qwen,liu2023visual,chen2024internvl}.
Given the unified processing of multiple modalities and rich pre-learned knowledge, MLLMs are increasingly applied to retrieval tasks, especially composed retrieval. Most approaches adapt the unified encoder directly with contrastive learning for embedding-based retrieval~\citep{jiang2024e5,zhang2024gme,huang2025joint,jiang2025vlmvec}, while others employ MLLMs as re-rankers for the later re-ranking~\citep{lin2024mm}. In this work, we study the embedding-based approaches.
An advantage of MLLMs for retrieval is the capability of jointly understanding contents in multiple modalities with a unified encoder. However, adopting the unified encoder in contrastive learning could also lead to shortcut learning~\citep{geirhos2020shortcut,wu-etal-2022-pcl}.
As a result, these recent methods generally inherit conventional contrastive loss and therefore remain vulnerable to modality shortcut when handling composed inputs.
Orthogonal to these previous works, we firstly study the modality shortcut problem in this setting and our proposed \ourmethod{} framework mitigates issue by introducing modality composition-aware objectives that can be easily integrated with MLLM-based retriever to enhance the robustness.
\section{Methodology}
\subsection{Preliminary: MLLM Embedding}
Multimodal retrieval aims to search for target documents by a given query. The embedding model, which is central to multimodal retrieval and the focus of our approach, is introduced in this section with the encoding process and training objectives.

\paragraph{Unified Multimodal Encoder}
Unlike conventional separate-encoder methods such as CLIP~\citep{radford2021learning, zhai2023sigmoid}, we study the approaches~\citep{lin2024mm, jiang2024e5, zhang2024gme, huang2025joint, jiang2025vlmvec} using a \emph{unified encoder} $f_\theta(\cdot)$, such as a MLLM. Formally, given an input $x$ that could comprises multiple modalities, $f_\theta(\cdot)$ encodes the input into a shared space as: $\mathbf{h} = f_\theta(x)$,
where $\mathbf{h}\in\mathbb{R}^d$ is the embedding in $d$ dimensions. In addition, $f_\theta(\cdot)$ includes the necessary processes such as tokenization and pooling, and we omit those details because it does not affect the modeling.

\paragraph{Contrastive Learning}
Then we can obtain the embedding of a query or document by the unified encoder. Prior works usually adopt the conventional CL loss to align the representations of queries and documents. To achieve this, we first define a function $\text{sim}_\theta(\cdot, \cdot)$ to measure the similarity between two inputs $x$ and $y$ as follows,
\begin{equation}
    \text{sim}_\theta(x, y) := \frac{\text{sim}[f_\theta(x), f_\theta(y)]}{\tau},
\end{equation}
where $\tau$ is the contrastive temperature parameter.
As $f_\theta(\cdot)$ is a unified encoder, $x$ and $y$ can be any modalities or the composition of them.
Next, the CL objective~\citep{radford2021learning} is optimized for aligning the query and positive document.
Given the dataset $\mathcal{D}$ that is constructed by queries and documents, in which each query is paired with its positive document, the CL loss can be formalized as follows,

\begin{equation}
\resizebox{.85\linewidth}{!}{$
    \mathcal{L}^{\text{CL}} = \mathbb{E}_{x,y^+,Y \sim \mathcal{D}} \left[-\log\frac{e^{\text{sim}_\theta(x,y^+)}}{e^{\text{sim}_\theta(x,y^+)}+\sum_{y \in Y}e^{\text{sim}_\theta(x, y)}}\right]
$}
\end{equation}

where $x$ is the query, $y^+$ is the paired positive document and $Y$ is the batch of negative documents.
By doing so, all examples are encoded into a shared space, within which the query is expected to be pulled closer to the positive document but pushed away from negative ones.

\begin{figure*}[ht]
    \centering
    \includegraphics[width=\textwidth]{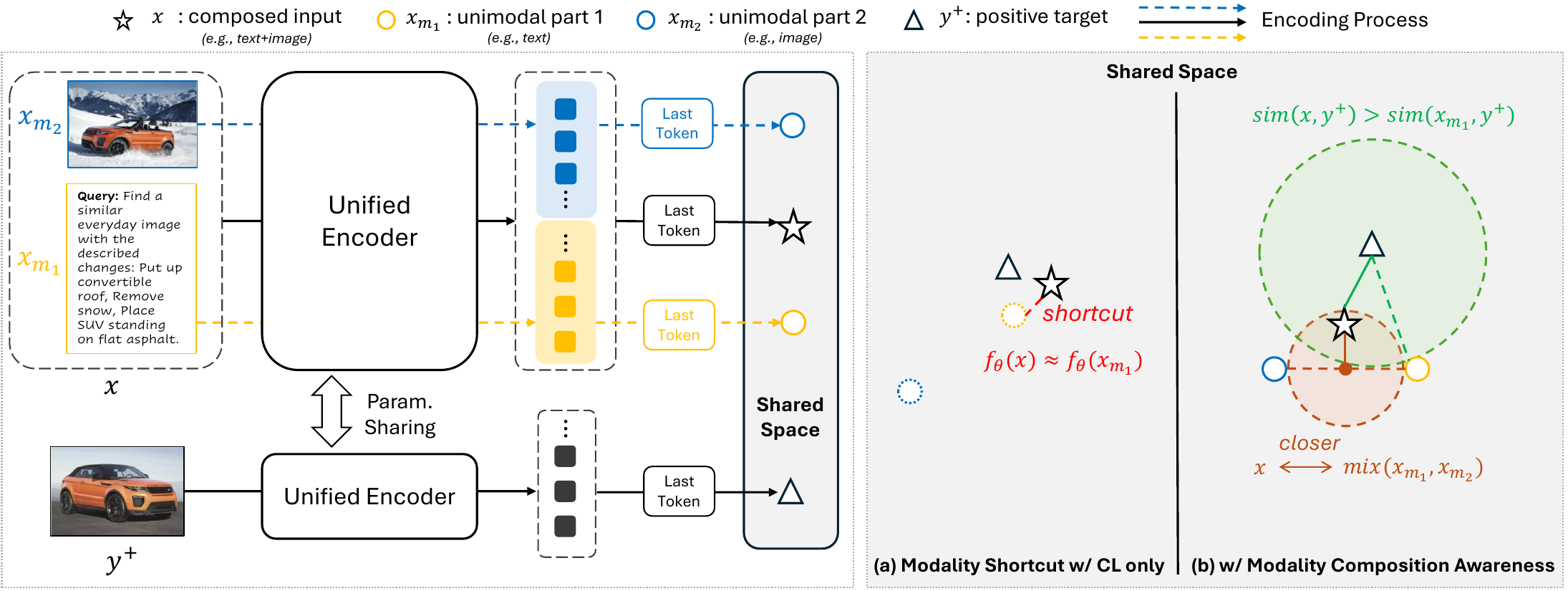}
    \caption{Modality Composition Awareness (\ourmethod{}, \S\ref{sec:ourmethod}). Unimodal parts are shown as dotted circles when not explicitly produced. \textbf{(a)} Training with vanilla CL: the composed embedding can be close to the target but still align disproportionately with one unimodal part, leading to modality shortcuts. \textbf{(b)} Training with CL and MCA: the composed embedding is explicitly constrained to be closer to the target than any of its unimodal parts by \textcolor{green}{MCP} (\S\ref{sec:mcp}), and anchored to a compositional prototype mixed (Equation~\ref{eq:mix}) from unimodal embeddings by \textcolor{brown}{MCR} (\S\ref{sec:mcr}) to mitigate modality shortcut.}
    \label{fig:approach}
\end{figure*}

\subsection{\ourmethod{}: Multimodal Embedding with Modality Composition Awareness}
\label{sec:ourmethod}
Although the conventional CL has demonstrated remarkable effectiveness in retrieval~\citep{radford2021learning,baldrati2022effective,liu2023universal}, it is inherently designed under the separate-encoder paradigm, where each modality is encoded independently and alignment is enforced at the representation level.
With the rapid development of MLLMs, the utilization of unified encoder for retrieval is becoming promising~\citep{jiang2024e5,zhang2024gme,huang2025joint,jiang2025vlmvec}.
However, those MLLM-based approaches still use conventional CL as the training objective, which is less adequate under composed scenario, where queries or documents may consist of multiple modalities.
Such approaches enable models to naturally handle composed inputs within a unified architecture, which could lead to modality shortcut learning as we introduced in \S\ref{sec:intro}.
Hence, a robust multimodal retrieval paradigm, particularly one that leverages MLLMs and adapts to a wider range of composed scenarios, requires not only aligning across modalities, but also explicitly modeling the \emph{modality composition}.

Consequently, we propose \emph{Modality Composition Awareness} to mitigate this issue. It consists of two components: (1) \emph{Modality Composition Preference} (MCP, \S\ref{sec:mcp}) and (2) \emph{Modality Composition Regularization} (MCR, \S\ref{sec:mcr}). These two training objectives are illustrated in Figure~\ref{fig:approach} and will be introduced in the following sections.
We first formalize the notation of modalities as $\mathcal{M}$\footnote{In this paper, we use text and image as available modalities, i.e., $\mathcal{M}=\{\text{text},\text{image}\}$, but the definition can be extended to more modalities.}.
Given a composed input $x$, we define the set of its unimodal counterparts as follows,
\begin{equation}
\label{eq:uniset}
    \mathcal{U}(x) := \{ x_m \mid m \in \mathcal{M}_x \},
\end{equation}
where $\mathcal{M}_x \subset \mathcal{M}$ is the modality set of $x$, and $x_m$ denotes its unimodal part for modality $m$. For example, if $x$ comprises text and image modalities, then $\mathcal{U}(x) = \{x_\text{text}, x_\text{image}\}$.

The modality shortcut problem can be formalized as $f_\theta(x) \approx f_\theta(x_{m})$ \emph{regardless of other modalities} after the optimization. In such a case the model only learns unimodal representation even composed inputs are provided. Formally, this key characteristic of modality shortcut is that the representation satisfies $f_\theta((x_m,x_\text{other})) \approx f_\theta((x_m,x'_\text{other}))$, meaning the model ignores variations in the other modality. As a result, when critical information resides in other modality under OOD conditions, the generalization performance could degrade.
Note that modality shortcuts only arise when at least one side of the input pair is a composed input. If both the query and document are unimodal, the task reduces to standard cross-modal retrieval with minimal shortcut risk. Therefore, the following proposed losses are only applied when either the query or document is composed.

\subsubsection{Modality Composition Preference}
\label{sec:mcp}
We propose a preference loss that explicitly enforces composed inputs to be more discriminative than their unimodal counterparts. Concretely, given a pair where either side is a composed input, we compute similarities between the composed embedding and candidate targets, as well as between its corresponding unimodal parts and the same target. The MCP loss is formalized as,
\begin{equation}
\label{eq:mcp}
\resizebox{.88\linewidth}{!}{$
    \mathcal{L}^{\text{MCP}} = \mathbb{E}_{x,y^+ \sim \mathcal{D}}
    \left[
    -\sum_{x_m \in \mathcal{U}(x)}\log 
    \frac{e^{\text{sim}_\theta(x,y^+)}}
    {e^{\text{sim}_\theta(x_{m},y^+)}}
    \right],
$}
\end{equation}
where $x$ and $y^+$ are the paired query-document sampled from dataset $\mathcal{D}$.
The MCP loss encourages the composed similarity higher than the unimodal similarities, ensuring that the model leverages complementary signals from multiple modalities rather than relying on one dominant modality. In other words, the loss formulates a preference-style constraint that whenever a composed input and its unimodal counterparts compete on the same retrieval task, the composed representation should be preferred.
For symmetry, we compute the loss in two directions, i.e., $x$ is the document and $y^+$ is the paired query, ensuring the preference is enforced for both query and document representation.

\subsubsection{Modality Composition Regularization}
\label{sec:mcr}
In addition to the preference constraint, we further propose MCR loss to encourage the consistency between the composed embedding and a simple prototype composition of its unimodal embeddings.
The intuition is that the representation produced by the encoder for a multimodal input should not deviate arbitrarily from the semantic space spanned by its unimodal parts.
To this end, we redefine the similarity function as follows,
\begin{equation}
\label{eq:mix}
\resizebox{.88\linewidth}{!}{$
    \text{sim}^{\text{mix}}_{\theta,\phi}(x, \mathcal{U}(x')) := \frac{\text{sim}[f_\theta(x), \text{mix}_\phi(\{{f_\theta(x'_{m}) \mid x'_{m}\in \mathcal{U}(x')}\})]}{\tau},
$}
\end{equation}
where $x$ and $x'$ are two inputs to compare. $x$ is the one we want to regularize, and $\mathcal{U}(x')$ denotes the set of unimodal components that we defined in Equation~\ref{eq:uniset}. $\text{mix}(\cdot)$ is a mixer module we defined that aggregates multiple unimodal embeddings into a composed prototype $\mathbf{h}'\in\mathbb{R}^d$ with the same output dimension as $f_\theta(\cdot)$.
Thus, the redefined similarity function measures the similarities between the composed prototype derived from $x'$ and the composed embedding of $x$.
In addition, to ensure that the regularization does not dominate learning, the mixer is deliberately kept simple, such as mean pooling or gated fusion, so that the encoder must learn to align composed inputs with the composed prototype rather than overfitting to the mixer.
\rev{
For example, we can instantiate the mixer as a modality-wise gated aggregation:}
\begin{equation}
\label{eq:gated}
\resizebox{.8\linewidth}{!}{$
\begin{aligned}
&[\mathbf{g}_1;\ldots;\mathbf{g}_M]
=
\mathrm{softmax}_{m}
\left(
\mathbf{W}[\mathbf{h}_1;\ldots;\mathbf{h}_M]+\mathbf{b}
\right), \\
&\mathrm{mix}_{\phi}^{\mathrm{gated}}(\{\mathbf{h}_m\}_{m=1}^{M})
=
\sum_{m=1}^{M}\mathbf{g}_m\odot\mathbf{h}_m .
\end{aligned}
$}
\end{equation}
\rev{
where $\mathbf{h}_m$ is the unimodal embedding of modality $m$, $\odot$ denotes element-wise multiplication, and $\mathrm{softmax}_{m}$ normalizes the gates across modalities for each embedding dimension.
}

The MCR loss is then defined as a contrastive objective, where the composed embedding is pulled closer to its mixed prototype than to other in-batch negatives,
\begin{equation}
\label{eq:mcr}
\resizebox{.88\linewidth}{!}{$
    \mathcal{L}^{\text{MCR}} = \mathbb{E}_{x,X\sim \mathcal{D}}
    \left[
    -\log\frac{e^{\text{sim}^{\text{mix}}_{\theta,\phi}(x,\mathcal{U}(x))}}
    {e^{\text{sim}^{\text{mix}}_{\theta,\phi}(x,\mathcal{U}(x))} + \sum_{x' \in X}e^{\text{sim}^{\text{mix}}_{\theta,\phi}(x, \mathcal{U}(x'))}}
    \right],
$}
\end{equation}
where $x$ is a sampled query or document, and $X$ is a batch of composed inputs that could be either queries or documents for symmetry.
Intuitively, the MCR loss enforces that the composed embedding remains anchored to the space formed by its constituent unimodal embeddings, thereby reducing the risk of spurious modality shortcut learning.
Finally, we combine the CL loss with the two proposed losses. The overall training objective for optimizing $\theta$ and $\phi$\footnote{$\phi$ is optimized only when the mixer contains learnable parameters.} is formulated as,
\begin{equation}
\label{eq:overallloss}
    \mathcal{L} = \mathcal{L}^{\text{CL}} + \alpha \times \mathcal{L}^{\text{MCP}} + \beta \times \mathcal{L}^{\text{MCR}},
\end{equation}
where $\alpha$ and $\beta$ are weighting coefficients controlling the relative strength of auxiliary terms. The impact of weighting is discussed in \S\ref{sec:ratio}.
\begin{table*}[ht]
\renewcommand{\arraystretch}{1.1}
\resizebox{\textwidth}{!}{
\begin{tabular}{l|*{8}{c}a|*{3}{c}a|*{4}{c}a}
\toprule
 & \multicolumn{9}{c|}{In-domain Retrieval} & \multicolumn{4}{c|}{OOD Retrieval} & \multicolumn{5}{c}{Zero-shot Grounding} \\
 \cline{2-19}
 Method
 & \rotatebox{90}{VisDial} 
 & \rotatebox{90}{CIRR} 
 & \rotatebox{90}{VisualNews\_t2i\quad} 
 & \rotatebox{90}{VisualNews\_i2t\quad} 
 & \rotatebox{90}{MSCOCO\_t2i} 
 & \rotatebox{90}{MSCOCO\_i2t} 
 & \rotatebox{90}{NIGHTS} 
 & \rotatebox{90}{WebQA}
 & \rotatebox{90}{\textbf{Avg.}}
 & \rotatebox{90}{OVEN} 
 & \rotatebox{90}{FashionIQ} 
 & \rotatebox{90}{EDIS} 
 & \rotatebox{90}{\textbf{Avg.}}
 & \rotatebox{90}{MSCOCO}
 & \rotatebox{90}{Visual7W-P} 
 & \rotatebox{90}{RefCOCO} 
 & \rotatebox{90}{RefCOCO-M}
 & \rotatebox{90}{\textbf{Avg.}} \\
\midrule
\multicolumn{19}{c}{\textit{Dual-Encoder Baselines}} \\
\midrule
CLIP~\citep{radford2021learning} & 30.7 & 12.6 & 78.9 & 79.6 & 59.5 & 57.7 & 60.4 & 67.5 & 55.8 & 41.1 & 11.4 & 81.0 & 44.5 & 33.8 & 55.1 & 56.9 & 61.3 & 51.7 \\
OpenCLIP~\citep{cherti2023reproducible} & 25.4 & 15.4 & 74.0 & 78.0 & 63.6 & 62.1 & 66.1 & 62.1 & 55.8 & 45.0 & 13.8 & 77.5 & 45.4 & 34.5 & 56.3 & 54.2 & 68.3 & 53.3 \\
SigLIP~\citep{zhai2023sigmoid} & 21.5 & 15.1 & 51.0 & 52.4 & 58.3 & 55.0 & 62.9 & 58.1 & 46.7 & 56.0 & 20.1 & 23.6 & 33.2 & 46.4 & 70.1 & 70.8 & 50.8 & 59.5 \\ 
BLIP2~\citep{li2023blip} & 18.0 & 9.8 & 48.1 & 13.5 & 53.7 & 20.3 & 56.5 & 55.4 & 39.5 & 39.3 & 9.3 & 54.4 & 34.4 & 28.9 & 52.0 & 47.4 & 59.5 & 46.9  \\
MagicLens~\citep{zhang2024magiclens} & 24.8 & 39.1 & 50.7 & 21.1 & 54.1 & 40.0 & 58.1 & 43.0 & 41.3 & 11.2 & 1.6 & 62.6 & 25.1 & 22.1 & 23.4 & 22.8 & 35.6 & 25.9 \\
\midrule
\multicolumn{19}{c}{\textit{Unified MLLM Encoder Baselines}} \\
\midrule
E5-V ~\citep{jiang2024e5} & 9.2 & 6.1 & 13.5 & 8.1& 20.7& 14.0& 4.2& 17.7 & 11.6 & 5.9 & 2.8 & 26.8 & 11.8 & 10.8 & 14.3 & 11.9 & 38.9 & 18.9  \\
VLM2Vec~\citep{jiang2025vlmvec} & 80.9 & 49.9 & 75.4 & 80.0 & 75.7 & 73.1 & 65.5 & 87.6 & 73.5 & 56.5 & 16.2 & 87.8 & 53.5 & - & - & - & - & -  \\
\midrule
\multicolumn{19}{c}{\textit{Fully-controlled Comparison}} \\
\midrule
$^\dagger$Qwen2-VL-2B ~\citep{wang2024qwen2}
& 81.3 & 51.6 & 75.7 & 78.6 & 73.9 & 71.9 & 65.6 & 86.8 & 73.2 & 63.0 & 15.6 & 84.0 & 54.2 & 41.0 & 60.7 & 60.3 & 69.6 & 57.9 \\

\ourmethod{} (Ours, 2B) & 80.3 & 50.5 & 76.3 & 78.7 & 74.6 & 72.1 & 66.1 & 86.8 & 73.2 \small{($\Delta\,0.0$)} & 66.7 & 19.3 & 86.1 & \textbf{57.4 \small{($\Delta+3.2$)}} & 42.7 & 65.4 & 65.0 & 70.9 & \textbf{61.0 \small{($\Delta+3.1$)}} \\

\bottomrule
\end{tabular}
}
\caption{Overall results.
$\dagger$ indicates the baselines reproduced with identical settings, backbone LLMs, and training data, and $\Delta$ indicates the difference compared to the completely comparable baselines.}
\label{tab:mainres}
\end{table*}
\section{Experiments}
\label{sec:exp}
To evaluate the proposed \ourmethod{}, we conduct extensive experiments on multimodal retrieval.
Since \ourmethod{} targets modality shortcut in composed retrieval, we focus on retrieval and grounding benchmarks where the target requires joint visual-textual constraints, and the details of  datasets and benchmarks are introduced in \S\ref{sec:datasets}. The training settings are introduced in \S\ref{sec:training}.
For evaluation we consider both IND benchmarks, which share the same distribution as the training data, and OOD benchmarks that test the generalization to unseen compositions, domains and tasks. OOD performance directly evaluates the shortcut learning problem~\citep{geirhos2020shortcut}, because the ideal solution generalizes to OOD test sets, but a shortcut-driven model may only performs well on training and IND test sets, which is illustrated in Figure~\ref{fig:oodfig}.
The OOD benchmarks cover two types of tasks: 1) \emph{OOD retrieval} that measures performance under distribution shifts such as domain and modality composition changes, and 2) \emph{OOD grounding} measures cross-task generalization. Both types of tasks require robust modality composition. Detailed evaluation settings are introduced in \S\ref{sec:evalsetting}.

\subsection{Main Results}
\paragraph{Overall Results.}
Our results in Table~\ref{tab:mainres} demonstrate the superior robustness of \ourmethod{} in OOD retrieval and zero-shot grounding scenarios.
\rev{
Note that some baselines inevitably differ in training data and backbone models.
For instance, VLM2Vec was explicitly trained on grounding tasks so it is not zero-shot performance.
To isolate the effect of modality composition awareness from other confounding factors,
we set up completely fair comparison between \ourmethod{} and our reproduced Qwen2-VL, which share identical experimental settings.
}
In general, MLLM-based methods outperform dual-encoder models because their unified encoder enables deeper cross-modal understanding, leading to better performance on all scenarios. However, MLLMs can suffer from modality shortcut issues, where the model relies too heavily on one modality as we introduced in \S\ref{sec:intro}. Our \ourmethod{} is designed to address this limitation and further enhance robustness.
Specifically, \ourmethod{} consistently achieves the highest average in OOD retrieval at 57.4, which is 3.2 points higher than the best baseline, showing a strong ability to generalize to unseen domains.
In zero-shot grounding, \ourmethod{} also leads with an average score of 61.0, outperforming the best baseline by 3.1 points, highlighting its advanced compositional understanding and effective transfer to tasks without explicit training. These results illustrate that modality composition awareness not only boosts robustness to distribution shifts but also enables the model to excel in zero-shot scenarios.
Importantly, \ourmethod{} maintains competitive in-domain retrieval performance, matching the best baseline at 73.2. This balance confirms that the gains in robustness do not come at the expense of accuracy on familiar data. \rev{More in-domain trade-off analysis is introduced in \S\ref{sec:in-domain-analysis}.}

\begin{figure}[ht]
    \centering
    \includegraphics[width=\linewidth]{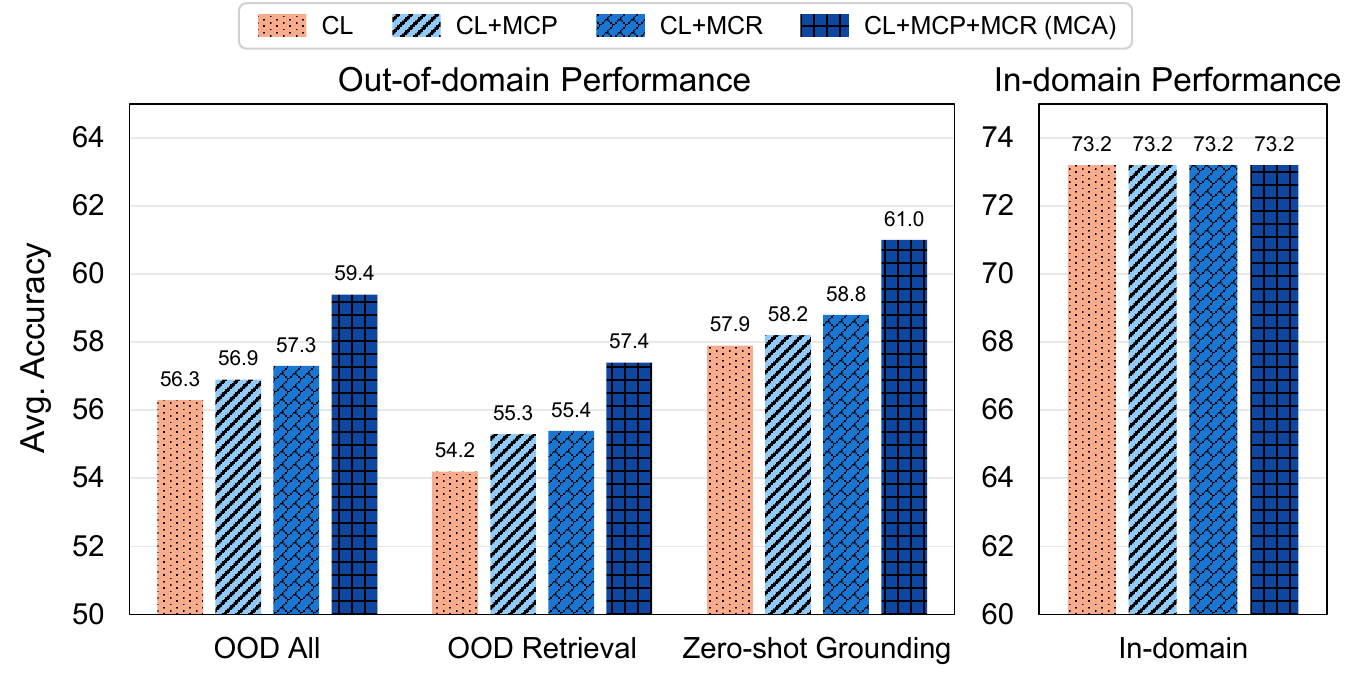}
    \caption{Breakdown of \ourmethod{} though loss ablation.}
    \label{fig:lossab}
\end{figure}

\paragraph{Loss Breakdown.}
As \ourmethod{} consists of two optimization targets, we further analyze the contribution of the two auxiliary losses separately.
As shown in Figure~\ref{fig:lossab}, both MCP and MCR individually lead to consistent positive gains on OOD benchmarks.
Notably, their standalone improvements are relatively smaller. When combined, the two losses yield substantially larger gains of +5.9\% on OOD retrieval and 5.4\% on zero-shot grounding, over the model trained with only contrastive learning, demonstrating the necessity of applying them together. This observation aligns with our design intuition in \S\ref{sec:ourmethod} that MCP encourages preference against unimodal shortcuts while MCR enforces structural consistency, and only their joint application forms the ideal constraint for robust modality composition, as illustrated in Figure~\ref{fig:approach}.
On IND benchmarks, all model variants converge to almost identical accuracy levels.
Note that the proposed \ourmethod{} introduces auxiliary losses to enhance model robustness. Therefore, similar in-domain performance compared to the baseline is expected and indicates that the conventional CL is sufficient to make the model well-converged on IND data. The main focus should be on improvements in OOD and zero-shot scenarios.
\rev{The full numbers of Figure\ref{fig:lossab} are listed in \S\ref{sec:full_breakdown}.}

\begin{figure*}[ht]
    \centering
    \begin{subfigure}[b]{0.24\textwidth}
        \centering
        \includegraphics[width=\textwidth]{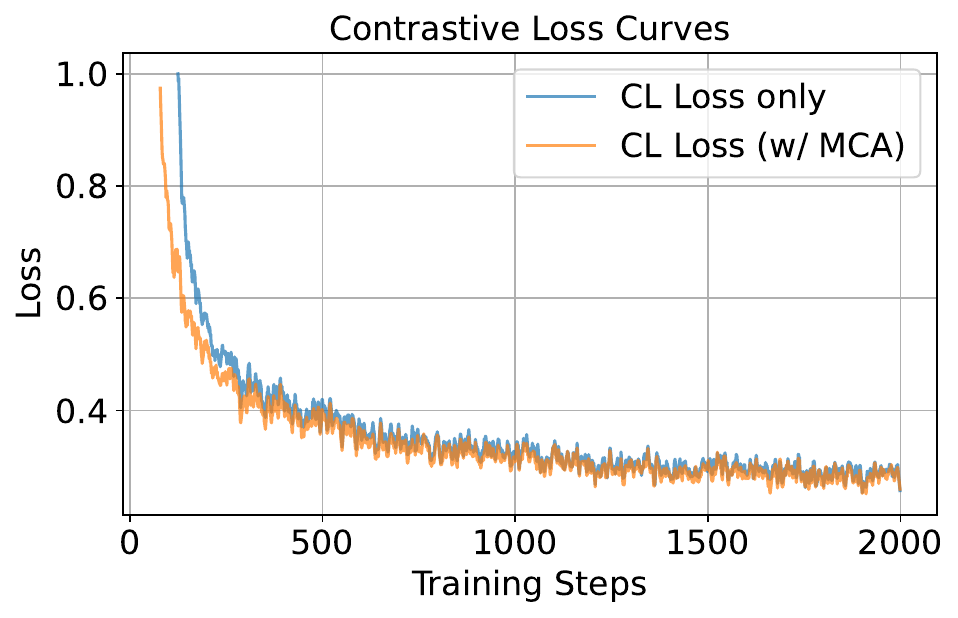}
        \caption{CL loss curve.}
        \label{fig:cl_loss_curve}
    \end{subfigure}
    \hfill
    \begin{subfigure}[b]{0.24\textwidth}
        \centering
        \includegraphics[width=\textwidth]{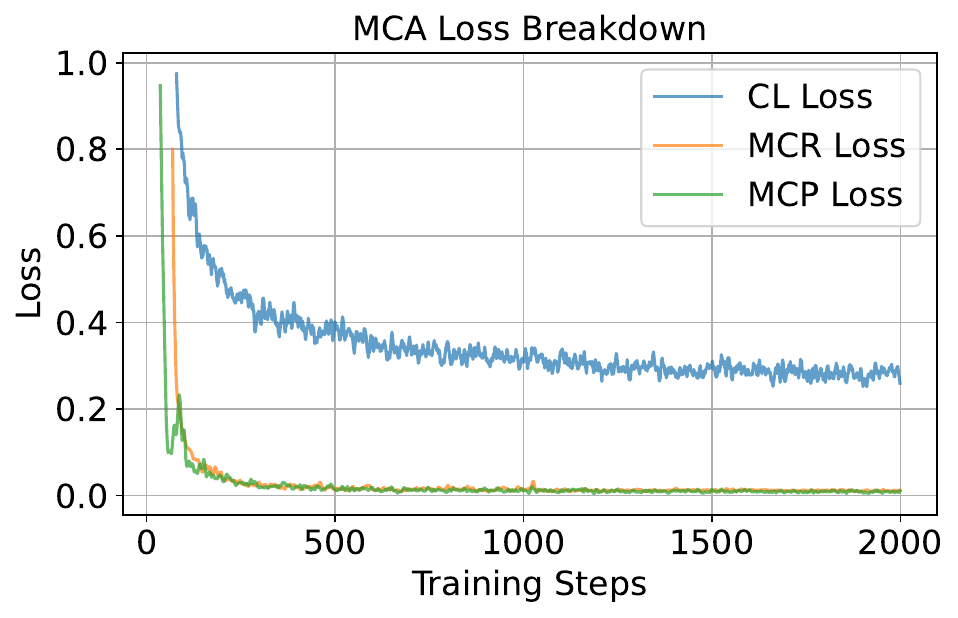}
        \caption{\ourmethod{} components loss.}
        \label{fig:mca_loss_curve}
    \end{subfigure}
    \begin{subfigure}[b]{0.24\textwidth}
        \centering
        \includegraphics[width=\textwidth]{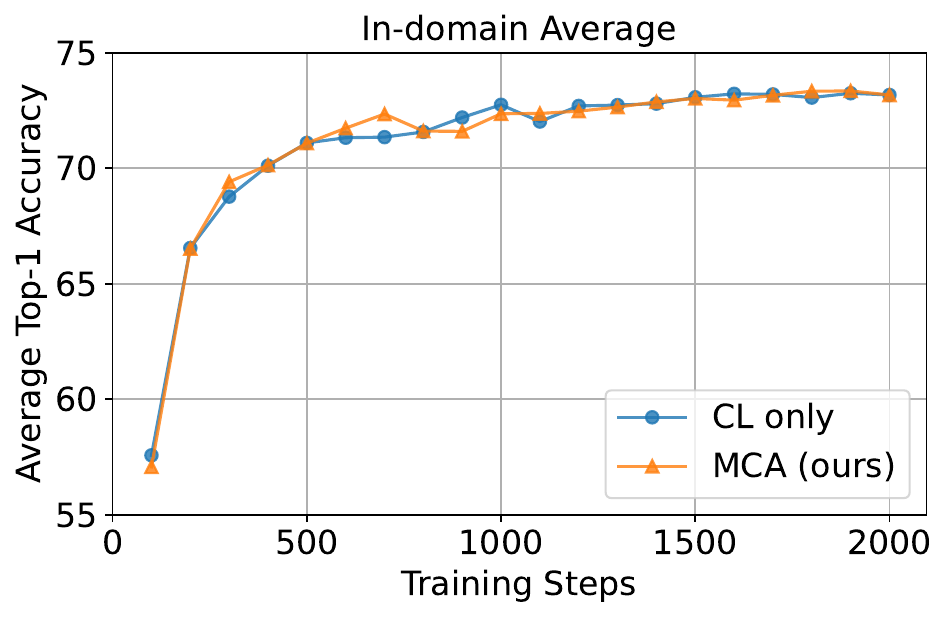}
        \caption{IND performance curve.}
        \label{fig:ind_curve}
    \end{subfigure}
    \hfill
    \begin{subfigure}[b]{0.24\textwidth}
        \centering
        \includegraphics[width=\textwidth]{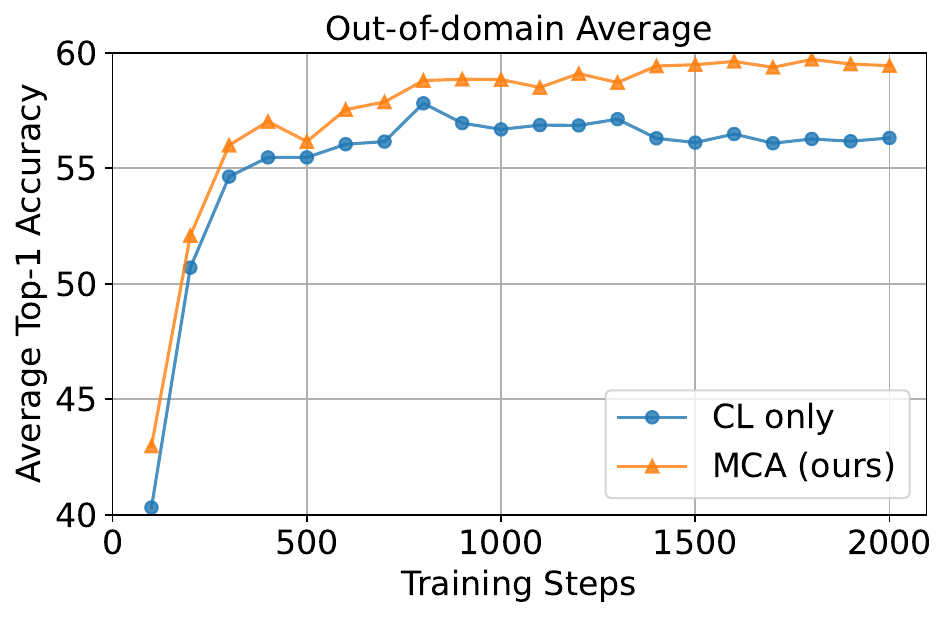}
        \caption{OOD performance curve.}
        \label{fig:OOD_curve}
    \end{subfigure}
    \caption{Convergence analysis of \ourmethod{}. For a fair comparison, the y-axis range for the loss is set to 0–1, and the range for benchmark performance is consistently set to 20.}
    \label{fig:curve}
\end{figure*}

\subsubsection{Convergence Analysis}
\label{sec:converge}

To assess the impact of the proposed \ourmethod{} loss on training dynamics, we perform a convergence analysis by comparing the loss curves and validation performance of both the baseline and the proposed method. This ensures that the improvements on OOD and zero-shot benchmarks are not caused by incomplete optimization or training instability.

\paragraph{Loss Curve}
We first examine whether the proposed objectives affect the optimization dynamics of the CL loss. Figure~\ref{fig:cl_loss_curve} compares the CL loss curves with and without \ourmethod{} during training. Both curves decrease smoothly and monotonically, indicating that introducing \ourmethod{} does not hinder the convergence of the main contrastive objective.
We then analyze the internal loss dynamics of \ourmethod{} in Figure~\ref{fig:mca_loss_curve}. The CL, MCP, and MCR losses all converge stably. We note that both MCP and MCR exhibit noticeable fluctuations at the beginning of training, possibly due to the unstable alignment between unimodal and multimodal embeddings in the early stage. Once the representation space becomes more coherent, both losses stabilize at small values.
\rev{
Notably, the MCP loss also serves as a dynamic measure of the modality shortcut gap, since it explicitly encourages the composed query to be more aligned with the target than its unimodal counterparts; therefore, its decrease and stabilization indicate that shortcut behavior is progressively suppressed during training.
}

\paragraph{Performance Curve}
To further track model performance during training, we measure the average retrieval accuracy every 100 training steps. Figure~\ref{fig:ind_curve} shows the IND results, where performance improves steadily and eventually converges to a level comparable to the baseline. In contrast, Figure~\ref{fig:OOD_curve} presents the OOD results, where \ourmethod{} consistently outperforms the baseline.
Moreover, the performance curves reveal that the benefits of \ourmethod{} appear after a few hundred steps. While IND accuracy converges similarly for models trained with and without \ourmethod{}, the OOD gap emerges within 500 steps and continues to widen throughout training. This suggests that \ourmethod{} enhances generalization by mitigating modality shortcuts. Overall, these results show that \ourmethod{} converges smoothly, with the auxiliary losses acting as regularizers, and leads to consistent improvements on OOD benchmarks where robustness is most critical.

\subsection{Input Resolution and \ourmethod{} Weighting}
\label{sec:ratio}

To better understand the behavior of \ourmethod{}, we investigate its sensitivity to two critical factors that are relevant: the resolution of image inputs and the weighting of proposed losses.
These two factors directly affect the balance between modalities. Specifically, when input information is degraded, the regularization strength might become crucial prevent shortcuts, while richer input information may require lighter regularization to avoid over-constraining the model.
Table~\ref{tab:ratio} summarizes the results across three input image resolution and different weighting values. For simplicity, we keep the coefficients of MCP and MCR loss equal, and vary their shared ratio to the CL loss.
\begin{table}[ht]
\centering
\resizebox{\linewidth}{!}{
\begin{tabular}{>{\centering\arraybackslash}m{1cm}|
                >{\centering\arraybackslash}m{1.8cm}|
                >{\centering\arraybackslash}m{1.8cm}
                >{\centering\arraybackslash}m{2cm}
                >{\centering\arraybackslash}m{1.8cm}}
\toprule
\ourmethod{} Ratio & IND Avg. & OOD Avg. & OOD Retrieval Avg. & Zero-shot Grounding Avg. \\
\midrule
\multicolumn{5}{c}{\textit{Low resolution (128 $\times$ 128)}} \\\hline
0 & 69.7 & 46.7 & 42.5 & 49.9 \\
0.01 & \cellcolor{red!1} 69.5 \small (-0.3\%) & \cellcolor{red!20} 44.0 \small (-5.8\%) & \cellcolor{red!30} 37.6 \small (-11.6\%) & \cellcolor{red!10} 48.8 \small (-2.2\%) \\
0.10 & \cellcolor{red!1} 69.3 \small (-0.6\%) & \cellcolor{blue!20} 49.5 \small (+5.9\%) & \cellcolor{blue!30} 46.8 \small (+10.1\%) & \cellcolor{blue!10} 51.4 \small (+3.0\%)\\
1.00 & \cellcolor{red!1} 69.5 \small (-0.3\%) & \cellcolor{blue!20} 50.9 \small (+8.9\%) & \cellcolor{blue!30} 48.1 \small (+13.1\%) & \cellcolor{blue!20} 52.9 \small (+6.0\%)\\
\hline
\multicolumn{5}{c}{\textit{Mid resolution (672 $\times$ 672)}} \\\hline
0 & 72.0 & 53.9 & 49.9 & 56.8 \\
0.01 & \cellcolor{red!1} 71.3 \small (-1.0\%) & \cellcolor{blue!10} 54.7 \small (+1.4\%) & \cellcolor{blue!20} 53.5 \small (+7.2\%) & \cellcolor{red!10} 55.5 \small (-2.3\%) \\
0.10 & \cellcolor{red!1} 71.5 \small (-0.7\%) & \cellcolor{blue!10} 54.9 \small (+1.8\%) & \cellcolor{blue!10} 51.3 \small (+2.8\%) & \cellcolor{blue!20} 57.6 \small (+1.4\%) \\
1.00 & \cellcolor{red!10} 70.9 \small (-1.6\%) & \cellcolor{blue!10} 55.8 \small (+3.5\%) & \cellcolor{blue!20} 53.9 \small (+8.0\%) & \cellcolor{blue!1} 57.3 \small (+0.8\%) \\
\hline
\multicolumn{5}{c}{\textit{High resolution (1344 $\times$ 1344)}} \\\hline
0 & 73.2 & 56.3 & 54.2 & 57.9 \\
0.01 & \cellcolor{blue!1} 73.2 \small (+0.0\%) & \cellcolor{blue!20} 59.4 \small (+5.5\%) & \cellcolor{blue!20} 57.4 \small (+5.9\%) & \cellcolor{blue!20} 61.0 \small (+5.4\%) \\
0.10 & \cellcolor{red!1} 73.0 \small (-0.3\%) & \cellcolor{blue!1} 56.6 \small (+0.5\%) & \cellcolor{red!10} 53.4 \small (-1.5\%) & \cellcolor{blue!10} 59.1 \small (+2.0\%) \\
1.00 & \cellcolor{blue!1} 73.2 \small (+0.0\%) & \cellcolor{blue!10} 58.3 \small (+3.5\%) & \cellcolor{blue!20} 57.0 \small (+5.1\%) & \cellcolor{blue!10} 59.3 \small (+2.4\%) \\
\bottomrule
\end{tabular}
}
\caption{Impact of weighting \ourmethod{} on varying input image resolution. Darker colors denote larger deviations from the vanilla CL.}
\label{tab:ratio}
\end{table}

Overall, we observe that higher resolution consistently yield stronger performance as expected, since richer visual detail provides more reliable visual grounding. Interestingly, the relative gain from \ourmethod{} grows larger as the resolution decreases. In lower resolution setting, where visual information is degraded and the risk of modality shortcuts is higher, \ourmethod{} provides significant improvements by enforcing the use of complementary textual cues. This patter suggest that \ourmethod{} is potentially more effective in scenarios with \emph{imbalanced modality} quality, where it helps the model resist collapsing onto a single modality and instead maintain faithful composition. Nevertheless, we also observe notable exceptions. In particular, under low-resolution inputs with very small weights 0.01, the model performs significantly worse than the vanilla CL baseline on OOD benchmarks. The possible reason is, the additional losses introduce small but inconsistent gradients, which are amplified  by the noisy low-resolution image representations. As a result, the model is shifted away from the vanilla CL optimum without receiving sufficient corrective signal, causing a collapse even more severe than the baseline.
In the mid-resolution setting, we also observe a modest degradation on IND benchmarks. this can be interpreted as the classic bias-variance trade-off. This degradation remains small, and the overall benefit of \ourmethod{} becomes more evident when robustness is prioritized.

In addition, the results generally highlight an interaction between input richness and the strength of \ourmethod{} regularization. When visual inputs are degraded, the risk of modality shortcuts is amplified, and stronger weighting is required for \ourmethod{} to be effective. In contrast, when inputs provide richer information, the model relies less on shortcuts and lighter weighting suffices to stabilize the training. This trend suggests a practical guideline: the richer the input information, the smaller the weighting can lead to a better generalization, whereas weaker inputs demand stronger regularization to preserve robust modality composition.

\subsection{Mixer Implementation}
\label{sec:mixer}
\begin{figure*}[ht]
    \centering
    \includegraphics[width=\linewidth]{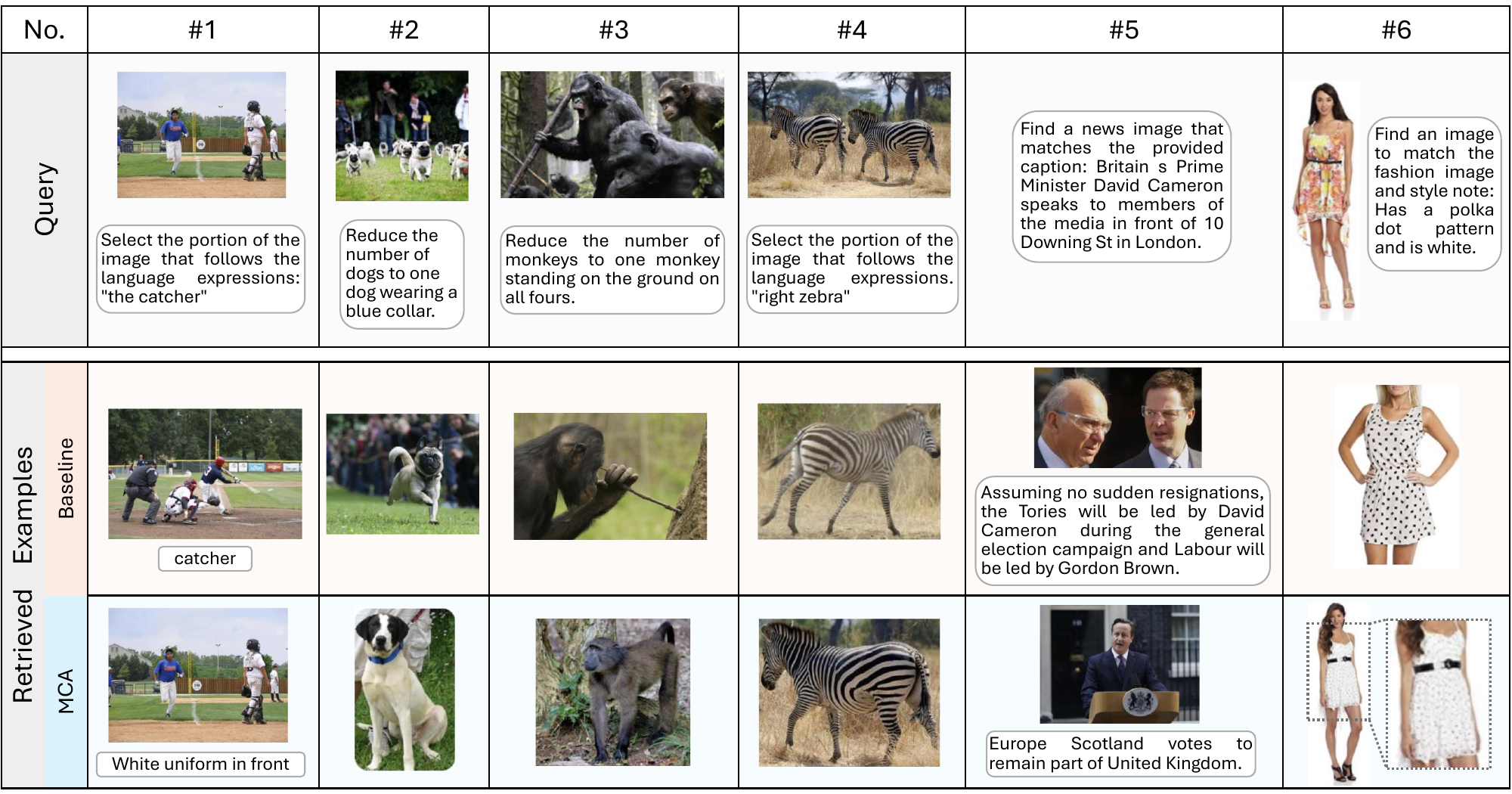}
    \caption{Qualitative examples. The central part of the retrieved image by \ourmethod{} for \#6 is zoomed in for a better visibility.}
    \label{fig:examples}
\end{figure*}

To examine the effect of different implementations of the mixer that defined in Equation~\ref{eq:mix}, we compare several simple modules, including the gated fusion module by default, the non-parametric mean pooling and the multimodal factorized bilinear pooling (MFB)~\citep{yu2017multi}. All mixer choices introduce only a negligible number of additional parameters, ensuring that the observed gains are not due to increased model capacity.
As shown in Table~\ref{tab:mixer}, the choice of mixer has little effect on IND benchmarks, suggesting that \ourmethod{} does not alter the standard retrieval behavior.
However, the improvements differ in OOD settings.
Both gated fusion and MFB variants yield improvements over the CL-only baseline, demonstrating the explicitly anchoring composed embeddings to the prototype from their unimodal parts is beneficial.
In contrast, the mean pooling provides only marginal improvements and even slightly degrades perfromance on cross-task grounding. This indicates that overly simple designs may be insufficient in practice to construct the prototype from unimodal embeddings.
Among them, the default gated fusion achieves the best performance. Above experiments suggest that the effectiveness of MCR does not heavily rely on the specific choice of the mixer, but rather on the general principle of enforcing compositional consistency, although the gain could vary depending on the mixer implementation.
\begin{table}[ht]
\centering
\renewcommand{\arraystretch}{1.1}
\resizebox{\linewidth}{!}{
\begin{tabular}{>{\centering\arraybackslash}m{2.2cm}|
                >{\centering\arraybackslash}m{1.8cm}|
                >{\centering\arraybackslash}m{1.8cm}
                >{\centering\arraybackslash}m{1.8cm}
                >{\centering\arraybackslash}m{1.8cm}}
\toprule
Mixer & In-domain Avg. & OOD Avg. & OOD Retrieval Avg. & Zero-shot Grounding Avg. \\
\midrule
Vanilla CL & 73.2 & 56.3 & 54.2 & 57.9  \\
\midrule
Mean Pooling            & \cellcolor{blue!1} 73.3 \small (+0.1\%) & \cellcolor{blue!1} 56.5 \small (+0.3\%) & \cellcolor{blue!10} 54.8 \small (+1.1\%) & \cellcolor{red!1} 57.8 \small (-0.2\%)  \\
Gated Fusion            & \cellcolor{blue!1} 73.2 \small (+0.0\%) & \cellcolor{blue!20} 59.4 \small (+5.5\%) & \cellcolor{blue!20} 57.4 \small (+5.9\%) & \cellcolor{blue!20} 61.0 \small (+5.4\%)  \\
MFB & \cellcolor{blue!1} 73.0 \small (-0.2\%) & \cellcolor{blue!10} 57.7 \small (+2.4\%) & \cellcolor{blue!1} 54.6 \small (+0.7\%) & \cellcolor{blue!10} 60.0 \small (+3.6\%) \\
\bottomrule
\end{tabular}
}
\caption{Impact of mixer. Darker colors denote larger deviations from the vanilla CL.}
\vspace{-13px}
\label{tab:mixer}
\end{table}

\subsection{Qualitative analysis}
\label{sec:examples}
Figure~\ref{fig:examples} presents qualitative comparisons between the baseline, i.e., training with only contrastive learning, and \ourmethod{} model, to highlight the behavior of modality shortcut. Across diverse cases from multiple benchmarks, the baseline fails by relying on partial cues in dominant modalities.
For example, in \#1 it select an irrelevant baseball scene because of the word \textit{``catcher''}, ignoring the visual grounding on the original image. In contrast, \ourmethod{} retrieves the image with textual description \textit{``white uniform in front''}. In \#2 and \#3, the baseline retrieves examples relying on image similarity only, but ignoring the textual instructions. \textit{``wearing a blue collar''} in \#2 and \textit{``standing on the ground on all fours''} in \#3 are not reflected, while \ourmethod{} grasps those information.
In \#4, the baseline mistakenly retrieves a zebra facing right, over-relying on the text cue and failing to ground the original image. \ourmethod{}, by contrast, localizes the \textit{``right zebra''} in the scene.
In the news retrieval case \#5, the query mentions \textit{``David Cameron speaking in front of 10 Downing St''}, yet the baseline retrieves the pair containing \textit{``David Cameron''} in both text and image but ignores the \textit{`` speaking in front of 10 Downing St''}. \ourmethod{} composes the texts and image thus finds the best match. Similarly, in the fashion example \#6, the baseline retrieves an image satisfying \textit{``polka dot and white''} but ignoring the original clothing style in the image. In contrast, \ourmethod{} integrates both modalities and selects the desired target.
Those examples demonstrate how \ourmethod{} mitigate modality shortcuts with the potential for more robust multimodal retrieval under various composed scenarios. In addition, we also present visualization of learned embeddings demonstrating that \ourmethod{} effectively reduce the modality shortcut in \S\ref{sec:tsne}. 

\section{Conclusion}
In this work, we introduced modality composition awareness (\ourmethod{}) for robust multimodal retrieval. By explicitly modeling the relationship between unimodal and multimodal inputs, \ourmethod{} incorporates two objectives: (1) modality composition preference, which discourages modality shortcuts by ensuring that composed representations are more discriminative than unimodal ones, and (2) modality composition regularization , which stabilizes embedding by aligning multimodal presentations with compositional prototypes. Extensive empirical experiments across IND and OOD benchmarks demonstrate that \ourmethod{} achieves consistent gains under distribution shifts and unseen zero-shot tasks, while maintaining comparable IND accuracy. These results highlight \ourmethod{} as an effective principle for mitigating modality shortcut problem when using MLLMs as unified encoder and improving generalization in multimodal retrieval.

\clearpage

\section*{Limitations}
While our proposed modality composition awareness show consistent improvements across various OOD benchmarks, several limitations remain.
First, the design of our losses assume that modality shortcuts primarily arise when composed inputs are present, and thus the framework is not directly applied to purely unimodal queries. This assumption may overlook shortcut phenomena that could still exist in separate-encoder approaches, although the risk is notably lower.
Second, the experimental results still show that our proposed \ourmethod{} could lead to performance degradation under particular weighting and input resolution setting. Although rare, this phenomenon suggests that it may requires setting adjust to make the proposed loss work in practice.
Third, in this work we only explore simple mixer implementation for fair comparison, and extending the design to richer mixers is an interesting open direction.
Lastly, although we have identified and studied the modality shortcut problem for vision-text, due to the lack of composed training data and benchmarks, the generalization to broader modalities, e.g., audio, has not yet been validated. In \S\ref{sec:gen_analysis} we present omni-modal generalization experiment but it is not thorough enough as most settings are empirically set.

\section*{Ethical Considerations}
This work focuses on improving multimodal representation learning by mitigating modality shortcut problem. Our experiments are conducted entirely on publicly available datasets and model checkpoints.
Our method does not directly generate new contents but instead focuses on retrieval, as a result, it poses minimal risk of misuse in generating harmful or deceptive contents.
Nevertheless, we emphasize the importance of responsible use of the retrieval model because the MLLMs could have unexpected biases.

\bibliography{custom}

\appendix
\clearpage
\section{Appendix}
\begin{figure}[ht]
    \centering
    \includegraphics[width=\linewidth]{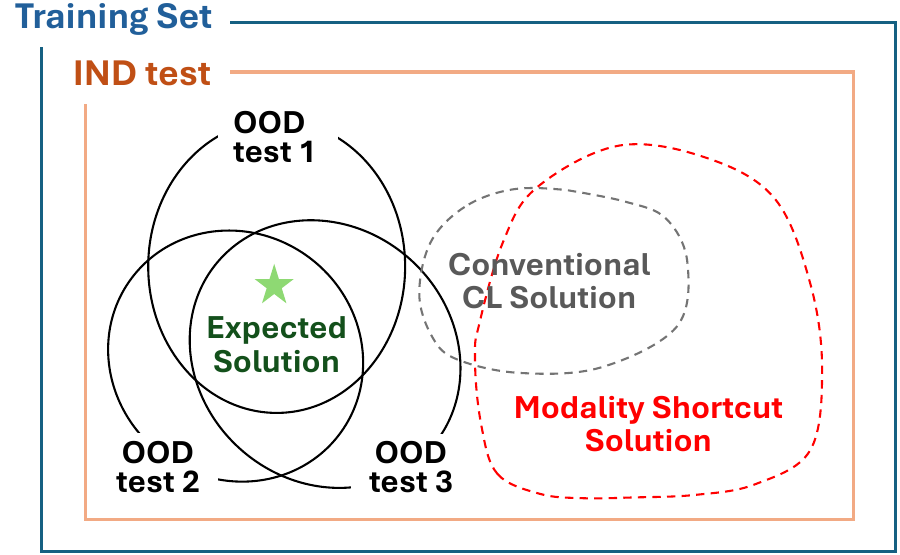}
    \caption{A conceptual diagram showing how OOD benchmarks evaluate modality shortcut.}
    \label{fig:oodfig}
\end{figure}

\subsection{Datasets}
\label{sec:datasets}
\paragraph{Training data.}
We utilize six multimodal retrieval datasets including both \emph{cross-modal} and \emph{composed} tasks for training: 
MSCOCO~\citep{lin2014microsoft},
VisualNews~\citep{liu2021visual},
VisDial~\citep{das2017visual},
CIRR~\citep{liu2021image},
NIGHTS~\citep{fu2023dreamsim}, and
WebQA~\citep{chang2022webqa}.
Table~\ref{tab:traindata} shows the number for examples and input modalities of training data.
We use the training splits from \texttt{MMEB}\footnote{\url{https://huggingface.co/datasets/TIGER-Lab/MMEB-train}} without additional curation. All models are trained jointly on the union of these datasets.
Although \ourmethod{} is designed for composed inputs, unimodal training pairs are also included. Each training batch contains a mix of unimodal and composed inputs. This ensures that the model learns basic representation ability for each modality, which is essential modeling of structural relationships among modalities that is required by our objectives.

\begin{table*}[ht]
\centering
\caption{Statistics of training datasets. I and T are the abbreviation of image and text, respectively.}
\renewcommand{\arraystretch}{1.4}
\resizebox{.9\textwidth}{!}{  
\begin{tabular}{l|cc|cc|c|c|c|c}
\toprule
Dataset & \multicolumn{2}{c|}{MSCOCO} & \multicolumn{2}{c|}{VisualNews} & VisDial & CIRR & NIGHTS & WebQA \\
\hline
Input modalities & \multicolumn{1}{c|}{I $\rightarrow$ T} & T $\rightarrow$ I & \multicolumn{1}{c|}{I $\rightarrow$ T} & T $\rightarrow$ I & T $\rightarrow$ I & (I+T) $\rightarrow$ I & I $\rightarrow$ I & T $\rightarrow$ (I+T) \\\hline
\# of training pairs & \multicolumn{1}{c|}{113K} & 100K & \multicolumn{1}{c|}{100K} & 100K & 123K & 26K & 16K & 17K \\
\bottomrule
\end{tabular}
}
\label{tab:traindata}
\end{table*}

\paragraph{Benchmarks.}
We evaluate the models on a range of multimodal retrieval benchmarks, which we group into \emph{IND} and \emph{out-of-domain} (OOD) settings.
The IND benchmarks correspond directly to the test splits of the six datasets used for training as mentioned in the previous paragraph and Table~\ref{tab:traindata}.
Since models that rely on modality shortcuts often exhibit poor generalization, we also assess this issue through robustness under OOD settings. The OOD benchmarks are deliberately selected:
OVEN~\citep{hu2023open} combines visual and textual modalities, where each instance consists of an image paired with a visual recognition question, alongside a reference Wikipedia image and its textual description including the title and first 100 tokens that serve as the target candidate for answering the question.
FashionIQ~\citep{hu2023open} features a multimodal composition of fashion product images paired with crowd-sourced textual descriptions that specify differences between products, enabling composed image retrieval where a reference image and modification text are combined to retrieve target images.
EDIS~\citep{liu2023edis} combines entity-rich text queries with multimodal candidates consisting of news images paired with their headlines, requiring models to understand both textual entities/events and visual content for retrieval.
MSCOCO-Grounding~\citep{lin2014microsoft,jiang2025vlmvec} transforms object detection into a multimodal ranking task where queries combine an image with a textual object name to retrieve cropped images of the specified object, with distractors sourced from other objects within the same image and from different images.
Visual7W-Pointing~\citep{zhu2016visual7w} combines textual questions with images to establish semantic links between descriptions and image regions, enabling both visual question answering through text-image composition and visual grounding through multimodal object localization.
RefCOCO~\citep{kazemzadeh2014referitgame} employs multimodal queries combining images with referring language expressions to identify specific objects, pairing image-text queries to retrieve cropped object images.
RefCOCO-Matching~\citep{kazemzadeh2014referitgame,jiang2025vlmvec} uses the same source datasets, while repurposed to match identical image-text compositions where both query and target contain the same object with its referring expression.
We directly use the test splits from \texttt{MMEB}\footnote{\url{https://huggingface.co/datasets/TIGER-Lab/MMEB-eval}}. Note that since modality shortcut in MLLMs, the problem we study in this paper, only happens in composed queries or documents, all of our OOD benchmarks focus on \emph{composed retrieval}, in which either query or document side comprise multiple modalities.

\subsection{Default training settings}
\label{sec:training}
By default, we adopt \texttt{Qwen2-VL-2B-Instruct}\footnote{\url{https://huggingface.co/Qwen/Qwen2-VL-2B-Instruct}} as the unified encoder backbone.
The contrastive temperature $\tau$ is set to $0.02$.
All models are fine-tuned with LoRA~\citep{hu2022lora} adapter. Unless otherwise specified, the LoRA rank is set to 8.
The overall loss combines conventional contrastive learning and our proposed MCR and MCP terms as introduced in Equation~\ref{eq:overallloss}, where $\alpha$ and $\beta$ are weighting hyper-parameters. These auxiliary terms act as regularization signals rather than main training objectives. We empirically adopt a small coefficient between 0.01 and 1.0 to stabilize training.
For simplicity, unless the otherwise specified, we tie $\alpha$ and $\beta$ by setting $\alpha = \beta = 0.01$ by default so that both the preference and regularization losses contribute equally.
Following previous work~\citep{jiang2025vlmvec}, we use AdamW optimizer with a learning rate of $2e-5$ and a linear schedule with warm-up. We also fixed the global batch size as $1024$ with gradient accumulation for varying numbers of GPU and memory. The total training steps are $2000$ and warm-up steps are $200$. The training takes around 200 hours on 8$\times$ Nvidia A100 40G.
For MCR loss, the mixer function $\text{Mix}_\phi$ in Equation~\ref{eq:mix} is instantiated as a simple gated fusion module by default, where each unimodal embedding is reweighted by a learnable gating coefficient before aggregation. For ablation, we also experiment with alternatives such as non-parametric mean pooling and multimodal factorized bilinear pooling~\citep{yu2017multi} in \S\ref{sec:mixer}.
All Qwen2-VL-7B results in \S\ref{sec:large_model} are reported at the final 2,000-step checkpoint, with MCA weights selected according to the resolution-dependent trend in Section~\ref{sec:ratio}: $\alpha=\beta=0.01$ for low resolution and $\alpha=\beta=1.0$ for mid resolution.

\subsection{Default Evaluation Settings}
\label{sec:evalsetting}
Unless otherwise specified, we report results using the final checkpoint of training for a fair comparison across all methods, as we have verified that the trend is stable across checkpoints in \S\ref{sec:converge}.
We report standard retrieval metrics accuracy@1 for all tasks.

\subsection{Embedding Visualization}
\begin{figure}
    \centering
        \begin{subfigure}[b]{0.23\textwidth}
            \centering
            \includegraphics[width=\textwidth]{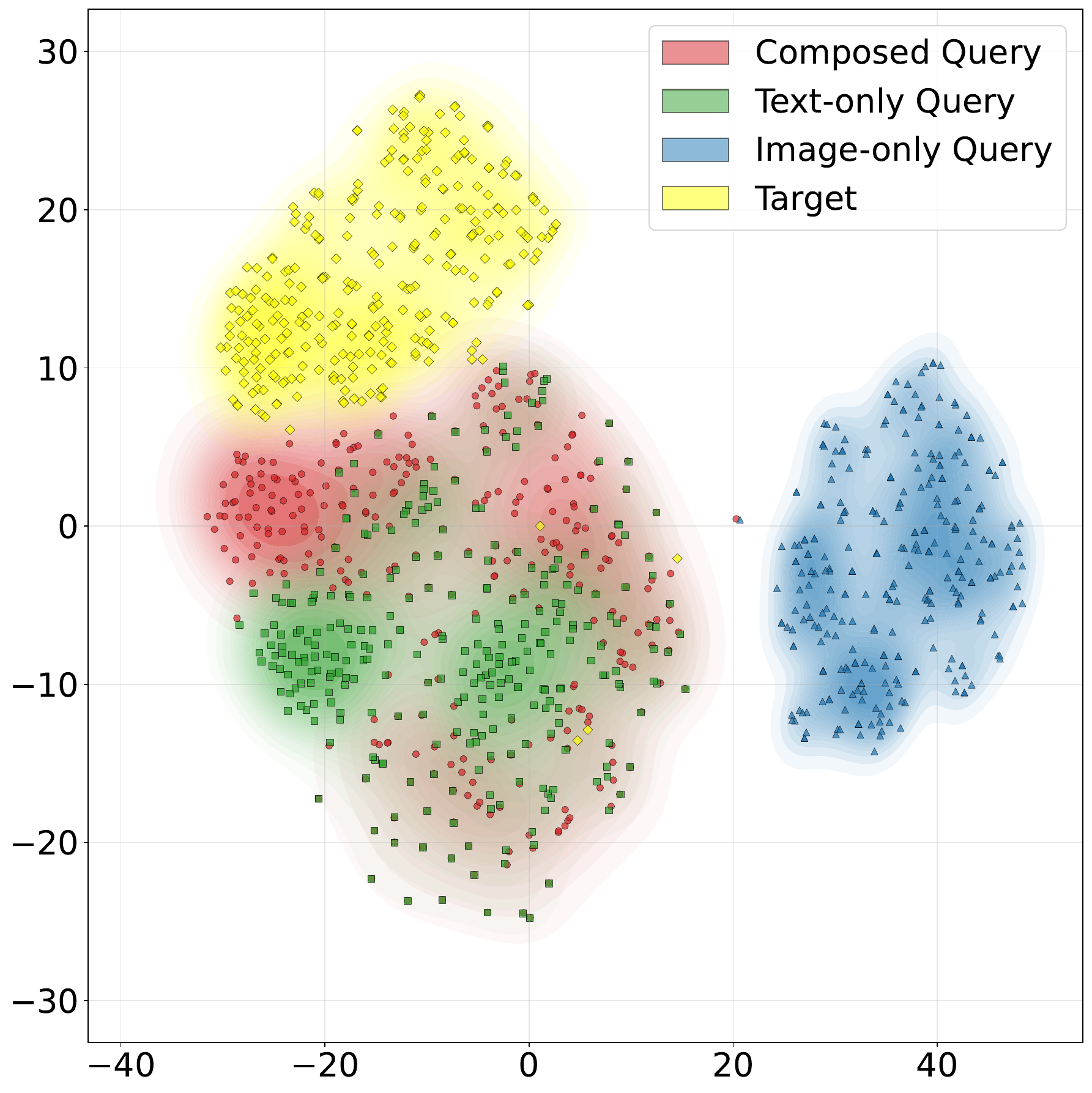}
            \caption{Trained with vanilla CL.}
            \label{fig:tsnebaseline}
        \end{subfigure}
        \hfill
        \begin{subfigure}[b]{0.23\textwidth}
            \centering
            \includegraphics[width=\textwidth]{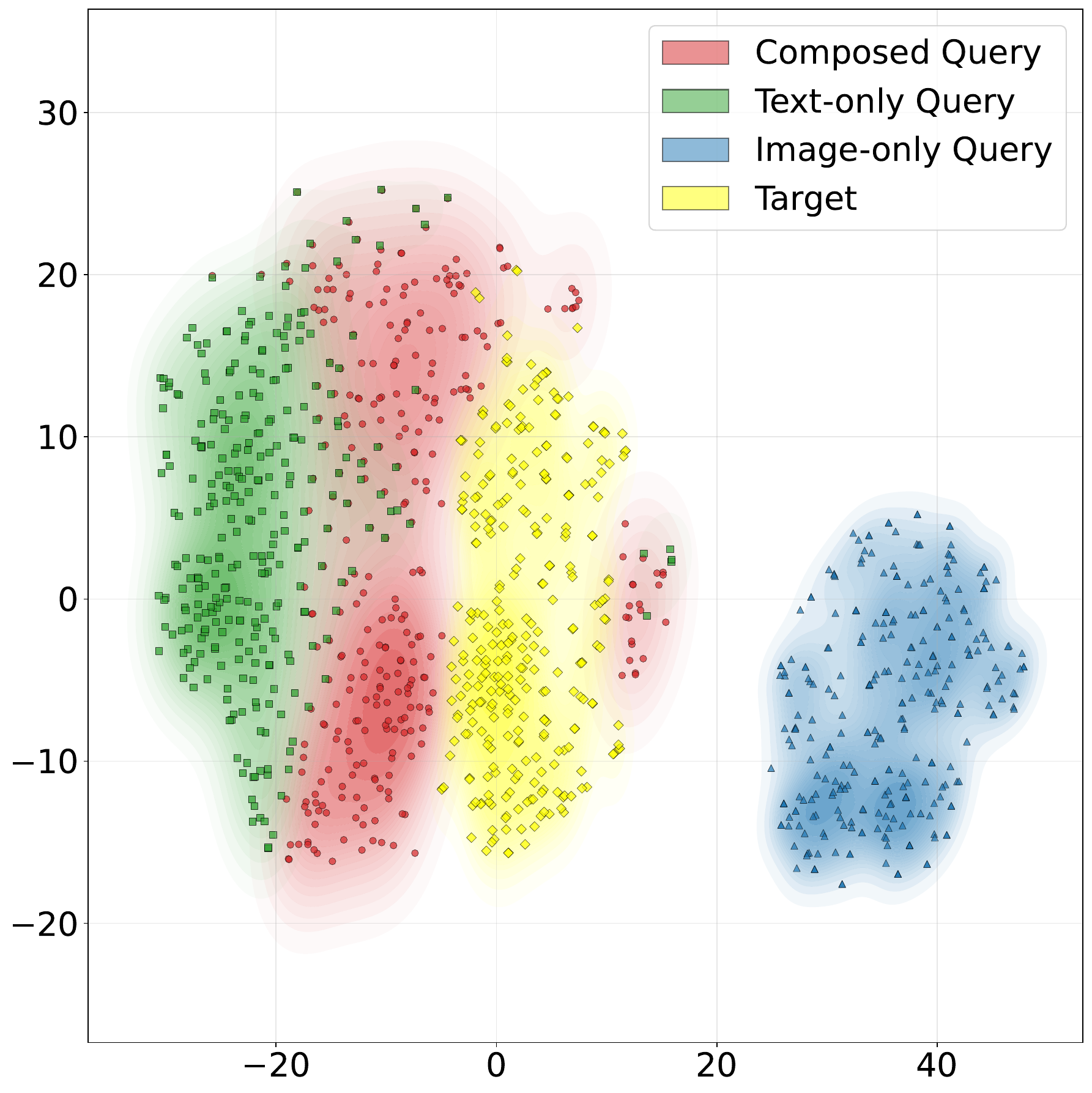}
            \caption{Trained with our \ourmethod{}.}
            \label{fig:tsneours}
        \end{subfigure}
    \caption{Visualizations of queries and targets.}
\end{figure}
In Figure~\ref{fig:tsnebaseline}, we visualize unimodal queries and targets randomly sampled from the CIRR dataset trained with vanilla CL and \ourmethod{}, respectively. Details of the t-SNE implementation are provided in \S\ref{sec:tsne}.
The visualization shows that, under vanilla CL, composed queries (red) are not well separated and frequently cluster close to text-only queries. This suggests that the model tends to exploit modality shortcuts rather than learning robust representations for composed queries. In contrast, with \ourmethod{} training, composed queries and text-only queries are clearly separated, indicating a more meaningful and robust representation distribution.

\subsection{Full Loss Ablation Results}
\label{sec:full_breakdown}
\begin{table*}[t]
\centering
\small
\setlength{\tabcolsep}{3.5pt}
\renewcommand{\arraystretch}{1.05}
\caption{
Per-dataset comparison of different training objectives.
}
\label{tab:per_dataset_breakdown}
\resizebox{\textwidth}{!}{
\begin{tabular}{l|cccccccc|c|ccc|c|cccc|c}
\toprule
\multirow{2}{*}{Method}
& \multicolumn{9}{c|}{In-domain Retrieval}
& \multicolumn{4}{c|}{OOD Retrieval}
& \multicolumn{5}{c}{Zero-shot Grounding} \\
\cmidrule(lr){2-10}
\cmidrule(lr){11-14}
\cmidrule(lr){15-19}
& VisDial & CIRR & VN-t2i & VN-i2t & COCO-t2i & COCO-i2t & NIGHTS & WebQA & Avg.
& OVEN & FashionIQ & EDIS & Avg.
& MSCOCO & Visual7W-P & RefCOCO & RefCOCO-M & Avg. \\
\midrule
Baseline
& 81.3 & 51.6 & 75.7 & 78.6 & 73.9 & 71.9 & 65.6 & 86.8 & 73.2
& 63.0 & 15.6 & 84.0 & 54.2
& 41.0 & 60.7 & 60.3 & 69.6 & 57.9 \\
CL+MCP
& 80.6 & 51.9 & 74.6 & 78.4 & 74.8 & 73.0 & 65.6 & 86.9 & 73.2
& 64.1 & 17.3 & 84.4 & 55.3
& 42.1 & 64.7 & 59.0 & 66.8 & 58.2 \\
CL+MCR
& 80.6 & 49.2 & 76.5 & 78.7 & 75.1 & 71.6 & 66.2 & 88.0 & 73.2
& 63.5 & 18.0 & 84.6 & 55.4
& 41.2 & 61.1 & 62.9 & 69.9 & 58.8 \\
MCA
& 80.3 & 50.5 & 76.3 & 78.7 & 74.6 & 72.1 & 66.1 & 86.8 & 73.2
& \textbf{66.7} & \textbf{19.3} & \textbf{86.1} & \textbf{57.4}
& \textbf{42.7} & \textbf{65.4} & \textbf{65.0} & \textbf{70.9} & \textbf{61.0} \\
\bottomrule
\end{tabular}
}
\end{table*}
\rev{
Table~\ref{tab:per_dataset_breakdown} provides a per-dataset breakdown of the results in Fig.\ref{fig:lossab}. 
All variants maintain nearly identical in-domain retrieval performance, with an average score of $73.2$, indicating that the proposed objectives do not sacrifice standard retrieval ability. 
In contrast, MCP and MCR provide complementary improvements on out-of-domain retrieval and zero-shot grounding. 
Combining them as MCA yields the strongest overall generalization, improving the OOD retrieval average from $54.2$ to $57.4$ and the zero-shot grounding average from $57.9$ to $61.0$. 
These results suggest that the two components encourage complementary aspects of multimodal composition, leading to more robust transfer beyond the in-domain retrieval setting.
}

\subsection{Generalization Analysis}
\label{sec:gen_analysis}
\begin{table}[ht]
\centering
\small
\setlength{\tabcolsep}{5pt}
\resizebox{\linewidth}{!}{
\begin{tabular}{lrrrr}
\toprule

& \textbf{\# Tasks}
& \textbf{Vanilla CL}
& \textbf{MCA}
& \textbf{$\Delta$} \\
\midrule
Overall
& 67 & 52.09 & \textbf{52.91} & +0.82 \\
Composed
& 28 & 55.78 & \textbf{56.96} & +1.18 \\
Non-composed
& 39 & 49.45 & \textbf{50.01} & +0.56 \\
\midrule
Image composed
& 18 & 61.90 & \textbf{62.81} & +0.91 \\
Video composed
& 8 & 43.87 & \textbf{45.59} & +1.72 \\
Audiovisual composed
& 2 & 48.33 & \textbf{49.80} & +1.47 \\
\midrule
Image non-composed
& 18 & 66.66 & \textbf{66.74} & +0.08 \\
Video non-composed
& 15 & 37.40 & \textbf{38.81} & +1.41 \\
Audio non-composed
& 6 & \textbf{27.92} & 27.82 & -0.10 \\
\bottomrule
\end{tabular}
}
\caption{
Omni-modal generalization results on 67 tasks. We report the macro-average Accuracy@1 across tasks. $\Delta$ denotes the absolute improvement in percentage points.
}
\label{tab:omni-generalization}
\end{table}
To better analyze the generalization of \ourmethod{} on more modalities and different backbone LLM, we conduct an experiment on \texttt{Qwen2.5-Omni-3B-Thinker} \footnote{\url{https://huggingface.co/Qwen/Qwen2.5-Omni-3B}}.
This experiment covers 67 tasks grouped according to their input modality and whether either side contains a modality composition. For brevity, we list the complete evaluation datasets
without repeating individual citations.
\textbf{Image composed:} A-OKVQA, CIRR, ChartQA, DocVQA, FashionIQ, GQA, InfographicsVQA, MSCOCO, N24News, OK-VQA, OVEN, RefCOCO, RefCOCO-Matching, ScienceQA, TextVQA, Visual7W, Visual7W-Pointing, and VizWiz. \textbf{Image non-composed:} Country211, EDIS, HatefulMemes, ImageNet-1K, ImageNet-A, ImageNet-R, MSCOCO-i2t, MSCOCO-t2i, NIGHTS, ObjectNet, Place365, SUN397, VOC2007, VisDial, VisualNews-i2t, VisualNews-t2i, WebQA, and Wiki-SS-NQ. \textbf{Video composed:} ActivityNetQA, Charades-STA, EgoSchema, MVBench, MomentSeeker, NExTQA, QVHighlight, and Video-MME. \textbf{Video non-composed:} Breakfast, DiDeMo, DiDeMo-V2T, HMDB51, K700, MSR-VTT, MSR-VTT-V2T, MSVD, MSVD-V2T, SmthSmthV2, UCF101, VATEX, VATEX-V2T, YouCook2, and YouCook2-V2T. \textbf{Audio non-composed:} AudioCaps, AudioCaps-A2T, Clotho, Clotho-A2T, MusicCaps, and MusicCaps-A2T. \textbf{Audiovisual composed:} VALOR32k-T2VA and VALOR32k-VA2T.

The comparison is fully controlled, following our default training setup, and both models are
fine-tuned with LoRA for 1000 steps with a global batch size 768. We set the LoRA rank to 16 and the
scaling factor to 32. We use last-token pooling,
L2-normalized embeddings, and a contrastive temperature of
0.02. For MCA, we set the weights of both MCP and MCR to 0.001.
For MCR, we instantiate $\operatorname{Mix}_{\phi}$ as
non-parametric mean pooling.
We report accuracy@1 for each task, and each aggregate result is computed as the unweighted macro-average of task-level accuracy@1 scores.
The gains still hold in this experiment, with larger improvements on composed tasks, which is consistent with MCA’s design. These results provide additional evidence that MCA can potentially scale to other unified encoder with broader modality coverage.

\subsection{Scalability to Model Size.}
\label{sec:large_model}
\begin{table}[t]
\centering
\small
\setlength{\tabcolsep}{5pt}
\resizebox{\linewidth}{!}{
\begin{tabular}{lcccc}
\toprule
\multirow{2}{*}{Mixer}
& In-domain
& OOD
& OOD Retrieval
& Zero-shot \\
& Avg.
& Avg.
& Avg.
& Grounding Avg. \\
\midrule
\multicolumn{5}{c}{\emph{Low resolution}} \\
\midrule
Qwen2-VL-7B
& 62.8
& 43.6
& 38.2
& 47.7 \\
MCA-7B
& \textbf{71.4} {\scriptsize (+8.6)}
& \textbf{55.2} {\scriptsize (+11.6)}
& \textbf{49.7} {\scriptsize (+11.5)}
& \textbf{59.3} {\scriptsize (+11.6)} \\
\midrule
\multicolumn{5}{c}{\emph{Mid resolution}} \\
\midrule
Qwen2-VL-7B
& 76.1
& 62.6
& 60.1
& 64.5 \\
MCA-7B
& \textbf{76.4} {\scriptsize (+0.3)}
& \textbf{63.6} {\scriptsize (+1.0)}
& \textbf{60.3} {\scriptsize (+0.2)}
& \textbf{66.1} {\scriptsize (+1.6)} \\
\bottomrule
\end{tabular}
}
\caption{
Category-level average results on Qwen2-VL-7B. OOD Avg. is computed over all seven OOD retrieval and zero-shot grounding tasks. Values in parentheses denote absolute improvements over the corresponding Qwen2-VL-7B baseline.
}
\label{tab:qwen2vl7b_avg}
\end{table}
\rev{
We further evaluate MCA on Qwen2-VL-7B\footnote{\url{https://huggingface.co/Qwen/Qwen2-VL-7B}} under both low- and mid-resolution settings, with detailed results reported in Table~\ref{tab:qwen2vl7b_avg}. Under low resolution, the 7B baseline shows clear robustness limitations, while MCA substantially improves all evaluation groups, with gains of +8.6 on in-domain retrieval, +11.5 on OOD retrieval, and +11.6 on zero-shot grounding. Under mid resolution, where the baseline is already much stronger, MCA still provides consistent improvements, especially on zero-shot grounding. This trend is consistent with the observations in \S~\ref{sec:ratio}. When visual inputs are degraded, modality imbalance becomes more severe and the model is more prone to shortcut learning, making composition-aware regularization more beneficial. These results show that MCA scales to larger backbones and is particularly effective when visual evidence is constrained.
}

\subsection{tSNE Implementation}
\label{sec:tsne}
In Figure~\ref{fig:tsnebaseline} and \ref{fig:tsneours}, we visualize the learned embedding spaces using t-distributed Stochastic Neighbor Embedding (t-SNE) to analyze the distribution characteristics of different query types. Embeddings were extracted from the CIRR dataset, which contains image-text compositions for retrieval tasks.
For dimensionality reduction, we employed t-SNE with perplexity=100, 1,000 iterations, and a random seed of 42 to ensure reproducibility. We randomly sampled 300 instances from each embedding type to maintain visual clarity while preserving distribution characteristics. To ensure fair comparison, we used identical sampling indices across all embedding types for baseline and our model.
The visualization combines kernel density estimation (KDE) with scatter plots to represent both the overall distribution and individual data points. We implemented a custom transparency gradient for the KDE plots, making low-density regions increasingly transparent to highlight areas of embedding concentration. Different marker shapes distinguish between embedding types: circles for composed queries, squares for text-only queries, triangles for image-only queries, and diamonds for target images.
The resulting visualization effectively captures the relationships between different modalities in the embedding space, revealing patterns of overlap and separation between composed, unimodal, and target representations. Comparing Figure~\ref{fig:tsnebaseline} and \ref{fig:tsneours}, it clearly indicates that vanilla CL could lead to a over-mixed distribution for composed and text-only embeddings, while \ourmethod{} learns a more robust representation with a clear boundary among different types of representations with a competitive IND performance and a better OOD performance.

\subsection{\rev{Training Efficiency}}
\rev{
MCA requires additional forward passes during training to compute unimodal embeddings for Equations~\ref{eq:mcp} and ~\ref{eq:mcr}. Since input sequence length satisfies $\sum_{m \in \mathcal{M}_x} L_m \approx L_{\text{composed}}$, the theoretical FLOPs overhead is approximately two times when text and image are included. Our measured wall-clock time shows 2.26× overhead on 8× A100 GPUs. This overhead can be further reduced because MLLMs are autoregressive, adn the intermediate hidden states for unimodal prefixes are already computed during the composed forward pass and can be reused to obtain unimodal embeddings without separate forward passes. We leave this implementation optimization for future work as it does not affect the model performance. Notably, MCA introduces no additional cost at inference time becasue the mixer is discarded after training.
}

\subsection{In-domain trade-off Analysis}
\label{sec:in-domain-analysis}
\rev{
We further analyze the slight per-dataset variations on in-domain benchmarks from a regularization--generalization perspective. Let $\mathcal{H}$ denote the hypothesis space under vanilla CL, and let $\mathcal{H}_{\text{MCA}} \subset \mathcal{H}$ denote the constrained space induced by MCA. For in-domain data where a shortcut solution $h_s \in \mathcal{H}$ achieves low loss, vanilla CL may converge to $h_s$, while MCA constrains the solution to $h \in \mathcal{H}_{\text{MCA}}$ that better respects modality composition structure. When such shortcuts are incidentally aligned with ground truth, $h_s$ may marginally outperform $h$ on certain in-domain tasks. However, under distribution shift where shortcuts become unreliable, $h$ is expected to generalize better. This reflects an inherent trade-off of regularization-based methods.
Specifically, MCA may slightly reduce in-domain flexibility, but it encourages more robust cross-modal composition. Empirically, the in-domain changes remain small, while MCA brings substantially larger gains on OOD retrieval and zero-shot grounding, aligning with its design goal of improving robustness for real-world deployment.
}

\subsection{Diagnostics of Modality Shortcut}
In addition to the OOD improvements, to directly verify whether the CL baseline exhibits modality shortcut and whether MCA reduces reliance on a single modality, we add a held-out shortcut diagnostic computed on composed test examples. For each composed input
$x$ paired with its positive target $y^+$, we define the composition
margin as
\begin{equation}
\resizebox{.88\linewidth}{!}{$
\Delta_{\mathrm{comp}}(x,y^+)
=
\operatorname{sim}\!\left[f_\theta(x),f_\theta(y^+)\right]
-
\max_{x_m\in\mathcal U(x)}
\operatorname{sim}\!\left[f_\theta(x_m),f_\theta(y^+)\right],
$}
\label{eq:composition-margin}
\end{equation}
where $\operatorname{sim}[\cdot,\cdot]$ denotes cosine similarity
between L2-normalized embeddings. A smaller margin indicates stronger shortcut tendency. We also report MCP loss defined in Equation~\ref{eq:mcp}, where a lower value
indicates a stronger preference for the composed representation. We compute both diagnostics on a fixed held-out probe containing 200
composed test examples, with 100 examples from CIRR and 100 examples
from NIGHTS. We use the same examples for the step-0 base model, the
final vanilla-CL checkpoint, and the final MCA checkpoint.

\begin{table}[ht]
\centering
\small
\setlength{\tabcolsep}{5pt}
\resizebox{\linewidth}{!}{
\begin{tabular}{lrrrr}
\toprule
\textbf{Model / checkpoint}
& \multicolumn{2}{c}{\textbf{Composition margin} $\uparrow$}
& \multicolumn{2}{c}{\textbf{Held-out MCP loss} $\downarrow$} \\
\cmidrule(lr){2-3}
\cmidrule(lr){4-5}
& \textbf{Value}
& \textbf{$\Delta$ vs.\ base}
& \textbf{Value}
& \textbf{$\Delta$ vs.\ base} \\
\midrule
Base model, step 0
& 0.1744
& -- 
& 0.0667
& -- \\
Vanilla CL, final
& 0.1248
& $-28.4\%$
& 0.0368
& $-44.8\%$ \\
MCA, final
& \textbf{0.2125}
& $\mathbf{+21.8\%}$
& \textbf{0.0100}
& $\mathbf{-85.0\%}$ \\
\bottomrule
\end{tabular}
}
\caption{
Held-out modality-shortcut diagnostics on 200 composed test examples
from CIRR and NIGHTS. A larger composition margin indicates clearer
separation from the strongest unimodal counterpart, while a lower
held-out MCP loss indicates a stronger preference for the composed
representation.
}
\label{tab:shortcut-diagnostics}
\end{table}

Table~\ref{tab:shortcut-diagnostics} shows that after conventional contrastive training, the composition margin decreases from 0.1744 to 0.1248. This means that although CL improves target matching, the composed query becomes less separated from its strongest unimodal counterpart, indicating stronger shortcut tendency. In contrast, MCA increases the final composition margin to 0.2125. We also observe that the composition-preference violation is much lower for MCA than for CL. Although CL also reduces this violation compared with the base model, MCA reduces it much more substantially, from 0.0368 to 0.0100. This shows that MCA more consistently assigns higher similarity to the composed query than to its unimodal counterparts.

\subsection{Failure Case Analysis}
\label{sec:fail}
\rev{
We observe a notable failure case in Table~\ref{tab:ratio}. Under low-resolution inputs (128×128) with small MCA weight, e.g., 0.01, performance degrades significantly compared to the baseline, specifically, -11.6\% on OOD Retrieval. We attribute this to the interaction between modality imbalance and regularization strength. At low resolution, visual representations become unreliable, creating a strong tendency toward text-only shortcuts, which can be a strong attractor in the optimization landscape. Small MCA weight provides insufficient corrective force to escape this attractor. Instead, it can perturb the optimization without guiding the model toward a composition-aware solution. In contrast, higher MCA weights successfully overcome the shortcut attractor by enforcing stronger compositional constraints, yielding improved OOD performance. Based on this analysis, in practice we recommend that when modality quality is imbalanced, for example, degraded images or noisy audio, stronger MCA weights are recommended to counteract shortcut tendencies, and when modalities are balanced and reliable, smaller weights are sufficient.
}

\subsection{Discussion on identity mapping}
\label{sec:identity_mapping}
\rev{
A potential concern of the proposed composition regularization is that it may admit a degenerate identity-mapping solution, where the composed representation $f_\theta(x_{\mathrm{comp}})$ simply collapses to the mixed unimodal prototype $\mathrm{mix}(f_\theta(x_{\mathrm{text}}), f_\theta(x_{\mathrm{image}}))$. In our formulation, however, the modality composition regularization is not used as a standalone objective, but as a soft regularizer jointly optimized with the retrieval-oriented contrastive and preference losses. Therefore, it only biases the representation toward compositional consistency rather than imposing a hard constraint. Moreover, such a collapse is unlikely in practice due to the strong capacity asymmetry between the MLLM encoder and the lightweight mixer. The encoder, initialized from a pretrained MLLM and updated with parameter-efficient fine-tuning, retains rich cross-modal reasoning capability, whereas the mixer only provides a low-capacity aggregation target. An identity-mapping solution would require the encoder to suppress its cross-modal computation and mimic this simple aggregation rule while still satisfying the retrieval objectives, which is disfavored by the constrained optimization. Thus, the regularization acts as a compositional anchor without dominating the learned representation.
}

\subsection{LLM Usage}
During the preparation of this paper, we utilized LLMs to assist with English proofreading. We also use LLM assistant for improving the efficiency of experiments such as organizing job execution scripts. Additionally, we used LLM assistant to support processing experimental results and generating figures.

\end{document}